%% file: ms.tex
\documentclass[letterpaper, 10 pt, conference]{ieeeconf}  

\IEEEoverridecommandlockouts                              

\usepackage{xcolor} 
\usepackage{amsmath}
\usepackage{algorithm} 
\usepackage{algpseudocode} 

\definecolor{darkgreen}{rgb}{0,0.5,0} 
\newcommand{\darkgreenbold}[1]{\textbf{\textcolor{darkgreen}{#1}}}
\definecolor{custompurple}{RGB}{128,0,128} 

\definecolor{darkorange}{RGB}{255,140,0}

\usepackage{graphicx}
\usepackage{amssymb}
\usepackage{booktabs}
\usepackage{afterpage}
\usepackage{lipsum} 
\usepackage{url}
\usepackage{hyperref}
\usepackage{booktabs}
\usepackage{caption}
\input{glossary}

\usepackage{cite} 

\title{\LARGE \bf
 Next Best Sense: Guiding Vision and Touch with FisherRF for 3D Gaussian Splatting} 

\author{Matthew Strong$^{\ast}{}^{1}$, Boshu Lei$^{\ast}{}^{2}$, Aiden Swann$^{3}$, Wen Jiang$^{2}$, Kostas Daniilidis$^{2}$, Monroe Kennedy III$^{3}$
\thanks{This research was supported by NSF Graduate Research Fellowship No.
DGE-2146755 and NSF Grant No. 2142773, 2220867.}
\thanks{$[\cdot]^{1} $Department of Computer Science,
$[\cdot]^{2}$School of Engineering and Applied Science, University of Pennsylvania, Philadelphia PA, USA. Emails: \{leiboshu, wenjiang, kostas\}@seas.upenn.edu.
$[\cdot]^{3}$Department of Mechanical Engineering,
        Stanford University, Stanford CA, USA.
        Emails: \{mastro1, swann, monroek\}@stanford.edu.
}
\thanks{$^\ast$Both authors contributed equally to this work.}
}

\begin{document}

\maketitle
\thispagestyle{empty}

\begin{abstract}
We propose a framework for active next best view and touch selection for robotic manipulators using 3D Gaussian Splatting (3DGS). 3DGS is emerging as a useful explicit 3D scene representation for robotics, as it has the ability to represent scenes in a both photorealistic and geometrically accurate manner. However, in real-world, online robotic scenes where the number of views is limited given efficiency requirements, random view selection for 3DGS becomes impractical as views are often overlapping and redundant. We address this issue by proposing an end-to-end online training and active view selection pipeline, which enhances the performance of 3DGS in few-view robotics settings. We first elevate the performance of few-shot 3DGS with a novel \textit{semantic depth alignment} method using Segment Anything Model 2 (SAM2) that we supplement with Pearson depth and surface normal loss to improve color and depth reconstruction of real-world scenes. We then extend FisherRF, a next-best-view selection method for 3DGS, to select views and touch poses based on depth uncertainty. We perform online view selection on a real robot system during live 3DGS training.  We motivate our improvements to few-shot GS scenes, and extend depth-based FisherRF to them, where we demonstrate both qualitative and quantitative improvements on challenging robot scenes. For more information, please see our project page at \href{https://arm.stanford.edu/next-best-sense}{arm.stanford.edu/next-best-sense}.
\end{abstract}




\section{INTRODUCTION}
\input{monroe_intro}

\section{PRELIMINARIES}
\label{sec:prelim}

\subsection{ 3D Gaussian Splatting}
3D Gaussian Splatting is an explicit 3D representation that represents a scene with 3D Gaussians. Each Gaussian is parameterized by a mean position $\mu \in \mathbb{R}^3$ and covariance $\Sigma \in \mathbb{R}^{3 \times 3}$. The covariance is computed as the product of a diagonal scale matrix $S \in \mathbb{R}^{3 \times 3}$ and rotation matrix $R \in \mathbb{R}^{3 \times 3}$ as $\Sigma = RS S^T R^T$. Each Gaussian also contains an opacity value $\alpha \in \mathbb{R}$ and spherical harmonic parameters to compute color. The color and depth of each pixel in a rendered image can be computed as blending ordered points along a camera ray, which with camera origin $\mathbf{o}$ and orientation $\mathbf{d}$ is defined as:
$\mathbf{r}(t) = \mathbf{o} + t\mathbf{d}$.
The color $C(\mathbf{r})$ and depth $D(\mathbf{r})$ is computed by blending ordered points intersecting the ray:
\begin{equation}
    \hat{C} = \sum_{i \in N} c_i \alpha_i T_i, \quad \hat{D} = \sum_{i \in N} d_i \alpha_i T_i,
\end{equation}

where $T_i = \prod_{j=1}^{i-1} (1 - \alpha_j)$, which is the transmittance, concretely defined as multiplying the previous Gaussian opacity values intersecting the ray. During training, the parameters of each Gaussian are optimized with gradient descent to minimize the following photometric loss betweeen a ground truth and rendered image:
as $\mathcal{L} = (1 - \lambda) \mathcal{L}_1 + \lambda \mathcal{L}_{\text{SSIM}}$, where $\mathcal{L}_1$ is the absolute loss between the RGB values of the rendered and ground truth image, and $\mathcal{L}_{\text{SSIM}}$ is the \textbf{S}tructural \textbf{S}imilarity \textbf{I}ndex \textbf{M}easure (SSIM) loss. Further, we can extend the loss to include $\mathcal{L}_{depth}$, which uses the ground truth and rendered depth image.

\subsection{Next Best View Selection}

 We use FisherRF \cite{jiang2023fisherrf} for next best view selection.  The problem of selecting the next best view is formulated as maximizing the Fisher information gain between candidate RGB camera poses $x_{i}^{\text{acq}}$ and views $y_{i}^{\text{acq}}$ and captured training RGB views $D^{\text{train}}$, given Gaussians' parameters $w^*$.

\begin{equation}
    \begin{aligned}
        &\mathcal{I}[w;\{ y_i^{\text{acq}}\}|\{ x_i^{\text{acq}}\}, D^{\text{train}}] \\
        & = H[w^*| D^{\text{train}}] - H[w^*|\{ y_i^{\text{acq}}\}, \{ x_i^{\text{acq}}\}, D^{\text{train}}]
    \end{aligned}
\end{equation}

Here, $H[\cdot]$ is the entropy. From \cite{BayesianActiveLearning}, the next best view objective is to maximize the reduction in entropy. In \cite{UnifyApproach}, the entropy of the model parameters can be approximated using a second order expansion.

\begin{equation}
    H[\omega^*] = -\frac{1}{2} \log \mathop{det} H''[w^*]
\end{equation}

Using the inequality $\log \det (A + Id) < \mathop{tr} (A)$, the final objective function for the next best view is \cite{UnifyApproach}:

\begin{equation}
    \mathop{\text{argmax}} \limits_{x_{i}^{\text{acq}}} \text{tr}(\mathbf{H}''[y_i^{\text{acq}} | x_i^{\text{acq}}, w^*]\mathbf{H}''[w^* | D^{\text{train}} ]^{-1})
\end{equation}

The Hessian matrix $\mathbf{H}''$ can be computed just from the Jacobian matrix, which only requires a single backward pass on a given view. In practice, we apply Laplace approximation that approximates the Hessian matrix with its diagonal values plus a prior regularizer \cite{jiang2023fisherrf,LaplaceRedux}.

\begin{equation}
\begin{aligned}
     \mathbf{H}''[y|x, w^*] &= \nabla_w f(x;w^*)^T \nabla_w f(x;w^*) \\
                   &\approx \text{diag}(\nabla_w f(x;w^*)^T \nabla_w f(x;w^*)) + \lambda I
\end{aligned}
\end{equation}

We present our extension to FisherRF for depth later in the methods section.

\section{METHOD}
\label{sec:method}
We now present \emph{Next Best Sense}. The method is divided into three  components: 1. \textbf{Few Shot Gaussian Splatting}, 2. \textbf{Next Best View for Robotics}, and 3. \textbf{Next Best Touch}. We leverage the real-time rendering capabilities of 3DGS to propose new views and touches in real-time.

\subsection{Few-Shot Gaussian Splatting}
In environments where robots are constrained to efficiently utilize a limited number of views, traditional Gaussian Splatting training collapses, suffering from poor reconstruction and overfitting. Specifically, \gls{3DGS} is known to be highly sensitive to initialization in the few-view case, requiring extra care to align the initialization points. 
We improve the performance of few-shot Gaussian Splatting by optimizing the manner in which depth is handled during initialization and training.

\begin{figure}[t]
    \centering
    \vspace{1.5mm}
    \includegraphics[width=0.95\linewidth]{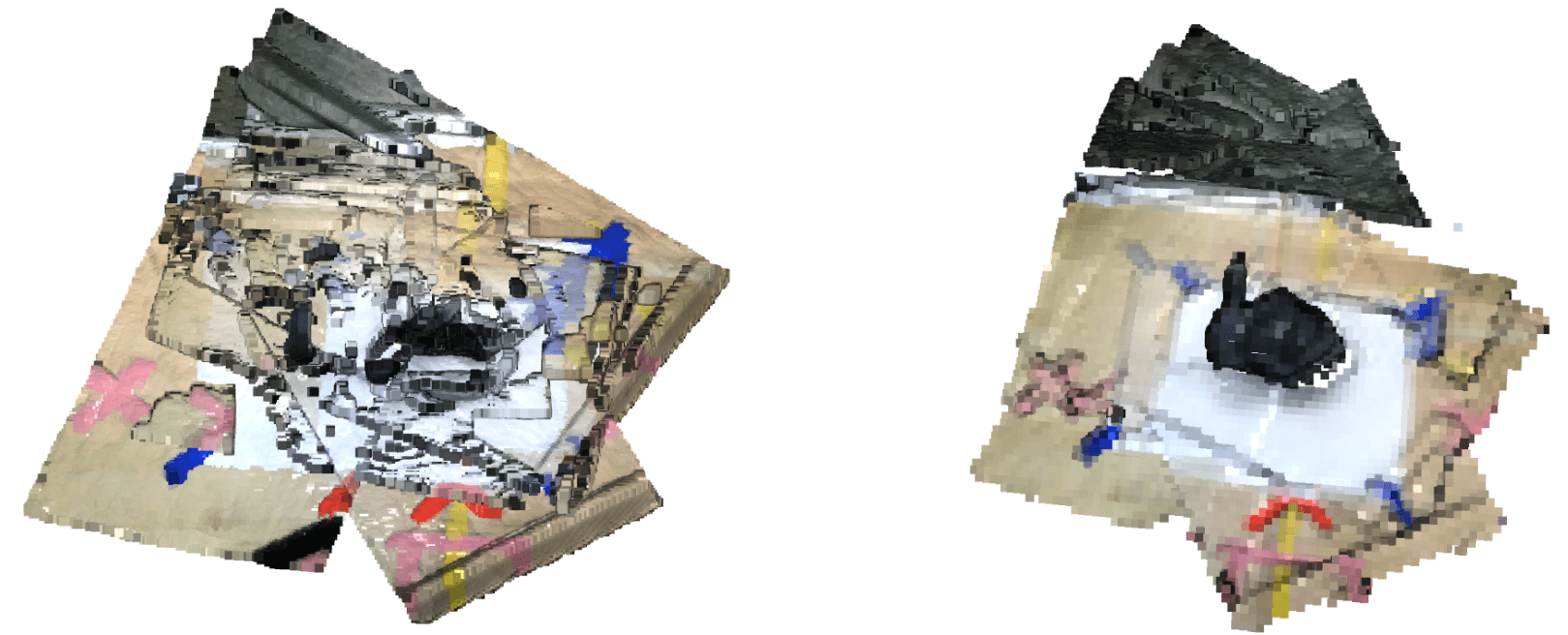}
    \caption{Mesh of Incorporating Lifted Depths (left) and Lifted \textbf{SAM2} Depths (right). A semantic alignment provides a robust initialization for 3DGS.}
    \label{fig:sam2-3d}
    \vspace{-2mm}
\end{figure}

\subsubsection{SAM2 Depth Alignment}

\begin{figure}[t]
    \centering
    \vspace{1.5mm}
    \includegraphics[width=0.99\linewidth]{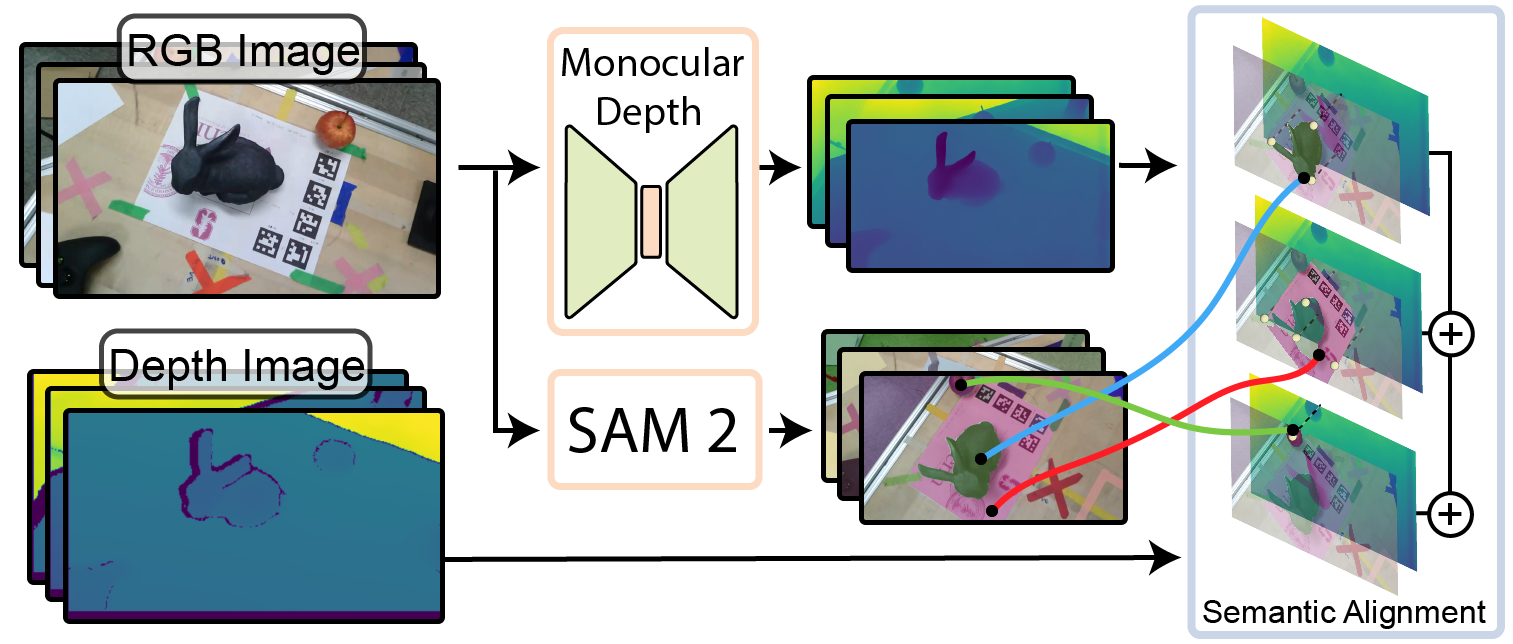}
    \caption{\textbf{SAM2 Alignment.} Given an RGB image and depth image, we provide the RGB as input to a monocular depth model to get relative depths, and run the SAM2 automatic mask generator to get object and scene masks. We then align each object in the monocular depth with the corresponding sensor depth.}
    \label{fig:semantic_alignment}
    \vspace{-3mm}
\end{figure}

We posit that depth alignment, which is done between a monocular depth image and real, noisy depth image, for Gaussian Splatting initialization, should be done \textit{semantically}, and supervision should be done \textit{relatively}. When aligning a real and monocular depth image and camera image, the depth of objects and components of a scene is based on the fact that the objects are semantically different. Concretely, we run the Segment Anything Model 2 automatic mask generator, which outputs a list of semantic masks $M$ in image $I$. The alignment can be observed in Fig. \ref{fig:semantic_alignment}. With an metric depth map from a depth sensor $D_{\text{real}}$, corresponding image $I$, monocular depth $D_{\text{MDE}}$, and list of masks $M$, we perform a mask-aware depth alignment:

\begin{equation}
s_{M_i}^*, t_{M_i}^* = \arg\min_{s,t} \sum_{p \in D_{\text{real}}} ||D_{\text{real}, M_i}(p) - D_{\text{MDE}, M_i}(p; s, t)||^2
\end{equation}
where $s_{M_i}^*$ is the scale factor on the pixels in mask $M_i$, $t_{M_i}^*$ is the offset, and $p$ is a depth keypoint from sensor depth image $D_{\text{real}}$. An example of the few shot depths back-projected into 3D and into a mesh can be found in Fig. \ref{fig:sam2-3d}. Finally, the user is asked to use SAM2's point prompt to mark the selected object in a single frame. We then propagate this to all other frames, which computes a mask. We add this mask to our list of masks to ensure object alignment. 

We \textbf{lift} the SAM2 aligned monocular depths into 3D and take a random percentage of them to initialize 3DGS, where a Gaussian of random color is constructed from each depth pixel. We find that this guides the few views to learn smoother geometries in the scene as compared to initialization from sensor depth data. Overall, this method maintains the metric accuracy of a depth sensor but the geometric quality of monocular depth models.

\subsubsection{Geometric Depth Guidance}
\label{gaussian-depth-guidance}
Drawing from prior works in few-shot GS \cite{zhu2023fsgs}, we introduce \textit{Pearson Relative Depth Loss} to gently guide depth in our scene. We use the monocular depth output from a model such as DepthAnythingV2 \cite{yang2024depth} or Metric3DV2 \cite{hu2024metric3d} and measure the distribution difference between rendered and monocular depth maps.

We also incorporate scale regularization in our work, detailed in PhysGaussian \cite{xie2023physgaussian}, to reduce highly anisotropic Gaussians in our scene. Normal regularization is also lightly used, in which we use work from DN-Splatter \cite{turkulainen2024dnsplatter} to derive normals, where we can approximate  the surface normals from the monocular depth map, and during training, we minimize one of the scaling coefficients during training to encourage the Gaussians to become a flat like disc. Finally, we apply camera optimization on our views, provided in \cite{nerfstudio} by adjusting the translation and rotation in $SO(3) \times \mathbb{R}^3$ for each view. 

\subsubsection{Learning a 3D SAM2 Object Mask}

We render a separate image from GS for our object mask, which we can then perform binary cross entropy with the SAM2 Mask. This allows us to compute the 3D Gaussians that are associated with the object, which we propose to use for touch.

\subsection{Next Best View for Robotics}

Real world robot environments are visually feature-rich and full of color. FisherRF will select views that have a high Hessian with respect to the rendered color image, meaning that candidate views with high gradients of color are more likely to be selected. We challenge this assumption that more colorful views are more uncertain. We assert that FisherRF should be extended to encode \textbf{depth uncertainty}, which is a strong indicator of geometric accuracy, and may be an even stronger than pure color uncertainty. To achieve this, we formulate a\textbf{ depth-based} negative log-likelihood objective  for the radiance field similar to \cite{jiang2023fisherrf}, which is simply the MSE between a ground truth and rasterized depth.
As this objective simply replaces color terms with depth terms in FisherRF, we follow the derivation in \cite{jiang2023fisherrf} and realize that depth information gain is nearly identical to the color information gain, but the outputs are different. We  define the information gain as a linear combination of the color and depth information gain ($\mathcal{I}_C$ and $\mathcal{I}_D$):

\begin{equation}
    \begin{aligned}
        &\mathcal{I}[w] =  \alpha \mathcal{I}_C[w; f_C] + \beta \mathcal{I}_D[w; f_D],
    \end{aligned}
    \label{eq:fisher-combo}
\end{equation}

where $\alpha$ weights the color information gain, and $\beta$ weights the depth information gain, and $f_C$ and $f_D$ are the color and depth rasterizers, respectively. Our formulation encodes the uncertainty as the depth gradient in a given view; discontinuous depths  and floaters are more likely to be removed. 


With the formulation of FisherRF, we perform our robotic procedure as follows.

\textbf{Collect Random Views.} First, $n$ ($n < 9$) random views are collected by sampling poses in a sphere of radius uniformly sampled from $r_{min} < r < r_{max}$, where the center camera ray points directly at the object center. For each view, we compute the monocular depth, SAM2 aligned monocular depth, and store these with the camera pose, color image, and SAM2  mask.

\textbf{View Selection During Training}: During training, the Gaussian Splat queries the robot for a list of feasible candidate views; we then run FisherRF to compute the next best view. The robot moves to the highest view score. If any errors are encountered in inverse kinematics or trajectory planning, we select the next highest scoring view.

\textbf{Add New View}: We then add the view, monocular depth, SAM2 aligned depth, camera pose, color image, and SAM2 object mask. We store the point prompt on the first view and propagate it to SAM2 as new views are added.

\begin{figure}[t]
    \centering
    \vspace{1.5mm}
    \includegraphics[width=0.95\linewidth]{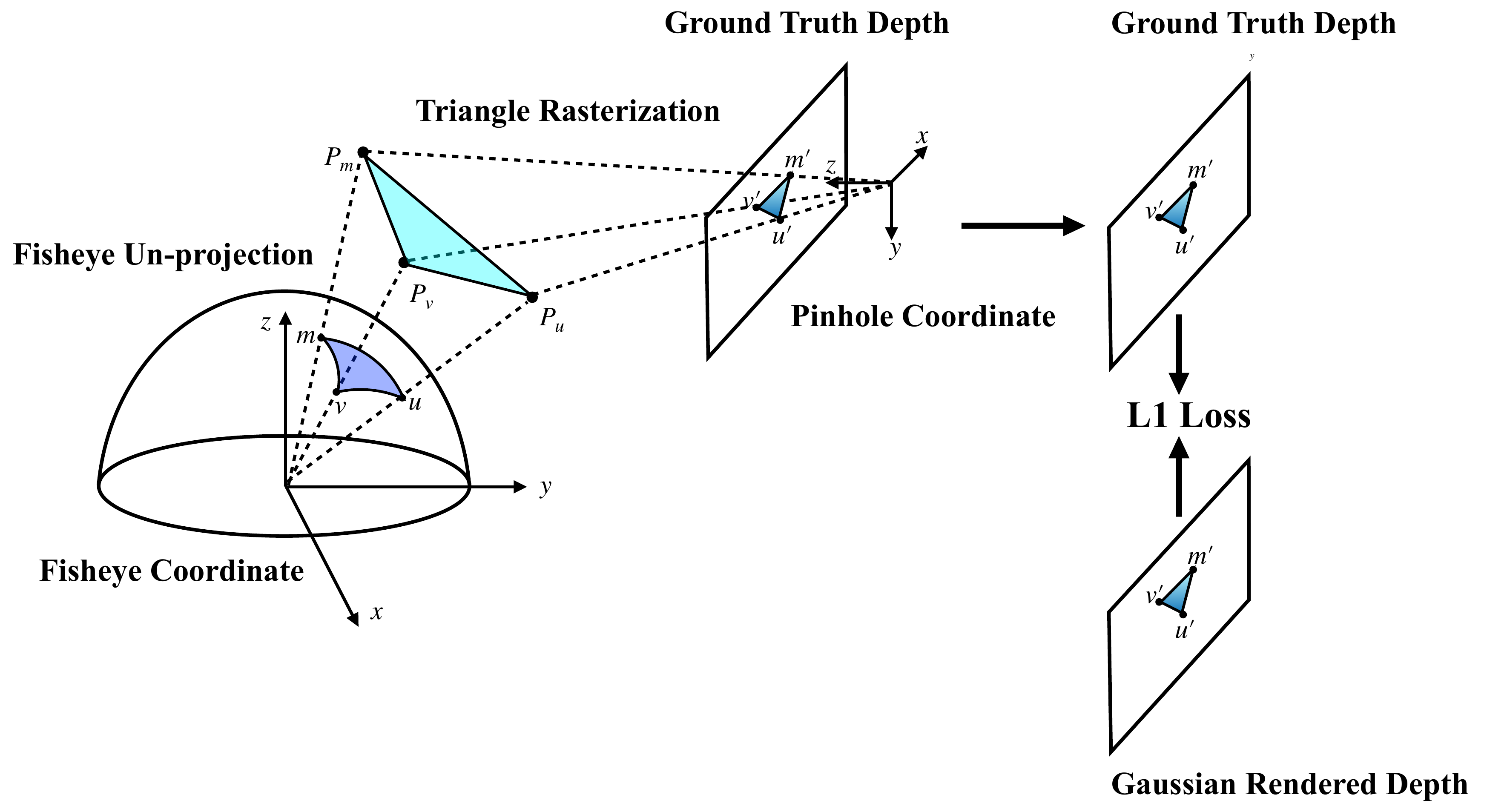}
    \caption{\textbf{Tactile Data Supervision}. We backproject points from a fisheye coordinate to form a triangle face $P_u, P_v, P_m$, which is projected onto image plane and rasterizes the corresponding area $u',v',m'$ for touch depth supervision.}
    \label{fig:tactile_supervision}
    \vspace{-7mm}
\end{figure}
\subsection{Next Best Touch}
Touch provides accurate local geometries of a surface,  regardless of lightning condition. It is then favorable to integrate it into our work as a tool to \textit{refine} a scene. To do this, we propose to use FisherRF depth after our vision phase to propose new touches for GS. 

We compute the next best touch by computing the expected information gain of a pseudo touch camera \textit{only on depth}. This ensures that more spikey and hole-like areas along an object are selected. We first only compute the depth uncertainty of Gaussians from the SAM2 object mask, ensuring spurious Gaussians not part of the object are ignored. As we add touches, we add them to the list of training views with different intrinsics to represent the touch camera. However, the touch poses are often too close to the surface of the object, which we address by moving the candidate touch view away from the object along its z axis till a surface of non-zero depth can be rendered.

We then collect a tactile image $I_t$ and its pose $p_i$ from a DenseTact \cite{do2022densetact} optical tactile sensor, where a fisheye camera fires rays to a soft hemispherical silicone surface. We use a pretrained DenseNet model \cite{DenseNet} to predict the depth along the ray from $I_t$, denoted as $I_d$. For each ray $\boldsymbol{t}_i$ from the tactile fish-eye camera, we unproject pixels into the point cloud point $\boldsymbol{p}_i = d \cdot \boldsymbol{t}_i$. We filter the points by their radial distance from the center and then by their z coordinate. Directly projecting the point cloud points to camera frame will lead to a sparse depth map, so we convert the point cloud to a triangle mesh using the connectivity from tactile image, and then rasterize the triangles into the image plane. The depth image is later used as depth supervision for Gaussian training, where we initialize Gaussians at the point of touch. We adopt $\mathcal{L}_1$ depth loss with smoothing around the point of touch, and the whole process is shown in Fig. \ref{fig:tactile_supervision}.

  \begin{table}[t]
    \vspace{5mm}
    \centering
    \begin{tabular}{lccc}
    \toprule
    Method & PSNR$\uparrow$ & SSIM$\uparrow$ & LPIPS$\downarrow$ \\
    \midrule
    \gls*{3DGS} w/o Depth  & 15.66 & 0.49 & 0.53 \\
    Dense-Depth \cite{chung2024depthregularized}  & {18.82} & {0.55} & {0.44}\\
    Pearson Loss Depth & 18.32 & 0.54 & 0.45  \\
    Lifted Depth, MSE Loss & 19.82 & 0.58 & \darkgreenbold{0.36} \\
    \textbf{Lifted SAM2 Depth, Pearson Loss} & \textbf{20.21} &\textbf{0.65} & \textbf{0.43} \\
    \textbf{Lifted SAM2 Depth, MSE Loss} & \textbf{20.33} & \textbf{0.67} & \textbf{0.41} \\
    \bottomrule
    \end{tabular}
    \caption{Blender few-shot scene ablations with \gls*{3DGS}}    
    \label{tab:ablation-results}
    \vspace{-4mm}
    \end{table}

\begin{table}[t]
    \vspace{1.5mm}
    \centering
    \begin{tabular}{lccc}
    \toprule
    Method & D-ABS$\downarrow$ & D-ABS-O$\downarrow$ \\
    \midrule
    Dense-Depth \cite{chung2024depthregularized}  & {0.27} & {0.20}\\
    Lifted Depth, MSE Loss & 0.18 & 0.035 \\
    \textbf{Lifted SAM2 Depth, MSE Loss} & \textbf{0.16} & \textbf{0.033}  \\
    \bottomrule
    \end{tabular}
    \caption{\textbf{Blender simulated few-shot} ground truth depth comparison. D-ABS (depth absolute error) is the error across the whole view; D-ABS-O is the error for only the bunny.}    
    \label{tab:depth-results}
    \vspace{-4mm}
\end{table}

\begin{figure}[t]
    \centering
    \vspace{3mm}
    \includegraphics[width=0.99\linewidth]{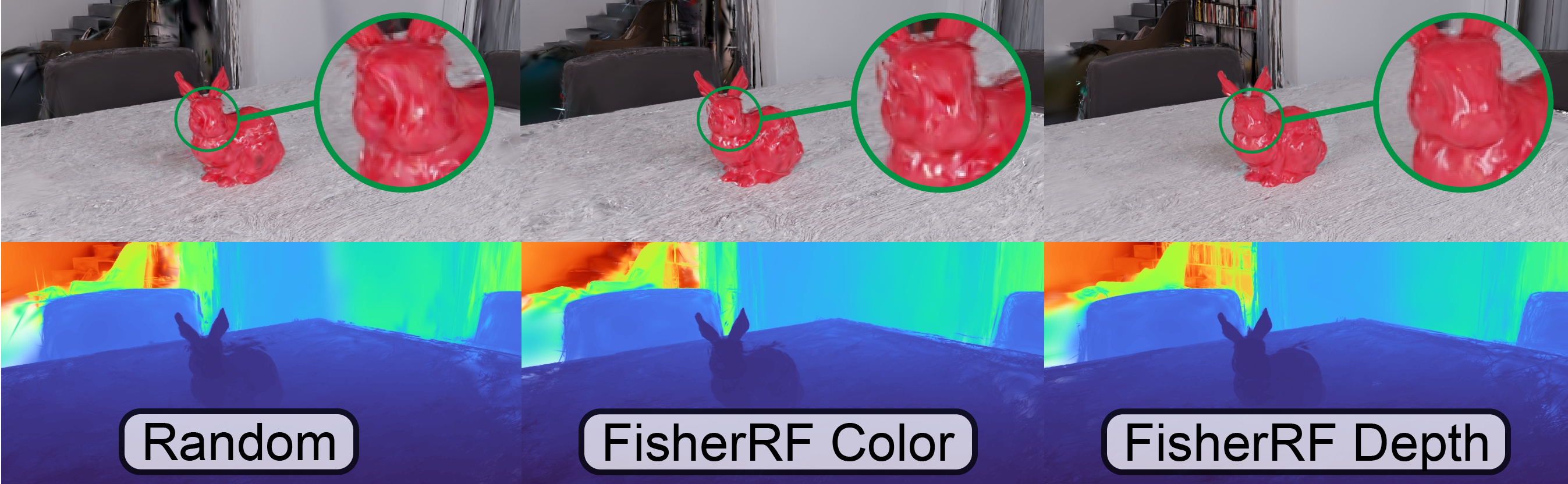}
    \caption{\textbf{FisherRF Ablations} Qualitative results of Random (Left), FisherRF Color (Middle) and FisherRF Depth (Right).}
    \label{fig:fisher-bunny}
\end{figure}

\begin{table}[ht]
    \vspace{3mm}
    \begin{center}
    \begin{tabular}{lccccc}
    \toprule
    Method & PSNR$\uparrow$ & SSIM$\uparrow$ & LPIPS$\downarrow$ \\
    \midrule
    3DGS  & 10.63 & 0.43 & 0.58 \\
    Dense Depth & 11.25 & 0.44 & 0.56 \\
    Pearson Loss Depth  & 11.68 & 0.47 & 0.54 \\
    Lifted Aligned Depth Pearson  & 11.79 & 0.47 & 0.53 \\
    SAM2 Lifted Aligned Depth Pearson & 11.84 & 0.47 & 0.54 \\
    Ours with Normal Supervision & 11.86 & 0.47 & 0.54 \\
   \darkgreenbold{Touch-GS \cite{swann2024touch}} &11.75 & 0.47 & \textbf{0.52} \\
   \textbf{Ours with Camera Optimization} &\textbf{12.30} & \textbf{0.50} & 0.54 \\
    \hline
    \end{tabular}
    \caption{Few-Shot Bunny Real Results}
    \label{tab:bunnyrealresults}
    \end{center}
\end{table}

\begin{table}[ht]
    \vspace{-2mm}
\begin{center}
\begin{tabular}{lcccccc}
\toprule
Object&Method & PSNR$\uparrow$ & SSIM$\uparrow$ & LPIPS$\downarrow$\hspace{-2mm} \\
\midrule
Mirror&Pearson-Depth  & 10.71 & 0.38 &  0.61 \\
&\textbf{Our Method}  & \textbf{10.74} & \textbf{0.40} & \textbf{0.59} \\
\hline
Prism&Dense-Depth  & 12.15 & \textbf{0.48} & 0.55 \\
&\textbf{Our Method}  & \textbf{12.24} & \textbf{0.48} & \textbf{0.53} \\
\hline
\end{tabular}
\caption{Few-Shot GS on Challenge Objects}
\label{tab:realworldchallengeresults}
\end{center}
\vspace{-4  mm}
\end{table}

\begin{table}[ht]
    \vspace{-2mm}
\begin{center}
\begin{tabular}{lcccccc}
\toprule
Object&Method & PSNR$\uparrow$ & SSIM$\uparrow$ & LPIPS$\downarrow$\hspace{-2mm} \\
\midrule
Bunny Blender&Random  & 23.68 & 0.68 &  0.24 \\
&FisherRF RGB  & 24.36 & 0.68 & 0.22 \\
&\textbf{FisherRF Depth}  & \textbf{24.64} & \textbf{0.69} & \textbf{0.21} \\
&\textbf{FisherRF Combined}  & \textbf{24.73} & \textbf{0.69} & 0.22 \\
\hline
Bunny Real&Random  & 12.20 & 0.47 & 0.49 \\
&FisherRF RGB  & 12.60 & \textbf{0.50} & \textbf{0.45} \\
&\textbf{FisherRF Depth}  & \textbf{12.67} & \textbf{0.50} & \textbf{0.45} \\
&\textbf{FisherRF Combined}  & \textbf{12.67} & \textbf{0.50} & \textbf{0.45} \\
\hline
\end{tabular}
\caption{Offline FisherRF}
\label{tab:realworldresultsoffline}
\end{center}
 \vspace{-5mm}
\end{table}

\begin{table}[t]
\begin{center}
\resizebox{0.48\textwidth}{!}{
    \begin{tabular}{lcccccc}
    \toprule
    Object&Method & PSNR$\uparrow$ & SSIM$\uparrow$ & LPIPS$\downarrow$\hspace{-2mm} \\
    \midrule
    &Random  & 12.25 & 0.51 & 0.54 \\
    Real-world&FisherRF RGB  & 12.49 & 0.52 & 0.56 \\
    Bunny&FisherRF Combined  & 12.43 & 0.52 & 0.56 \\
    &\textbf{FisherRF Depth}  & \textbf{13.09} & \textbf{0.53} & \textbf{0.53} \\
    \hline
    Table Scene&Random  & 13.39 & \textbf{0.65} & 0.55 \\
    Bunny+Ridged Object&FisherRF RGB  & 13.47 & \textbf{0.65} & 0.55 \\
    &\textbf{FisherRF Depth}  & \textbf{13.67} & \textbf{0.65} & \textbf{0.54} \\
    \hline
    Toy Car&Random  & 11.40 & 0.45 & 0.55 \\
    &\textbf{FisherRF Depth}  & \textbf{11.75} & \textbf{0.47} & \textbf{0.52} \\
    \hline
    Mirror&Random  & 14.66 & 0.72 & 0.50 \\
    &\textbf{FisherRF Depth}  & \textbf{16.50} & \textbf{0.76} & \textbf{0.41} \\
    \hline
    Prism&Random  & 16.63 & 0.73 & 0.41 \\
    &\textbf{FisherRF Depth}  & \textbf{16.77} & \textbf{0.74} & \textbf{0.39} \\
    \hline
    \end{tabular}
}
\caption{FisherRF Online Real-World Experiment}
\label{tab:realworldresults}
\end{center}
\vspace{-4mm}
\end{table}

\section{RESULTS}
\label{sec:results}

\begin{figure*}
    \centering
    \vspace{3mm}
    \includegraphics[width=0.55\linewidth]{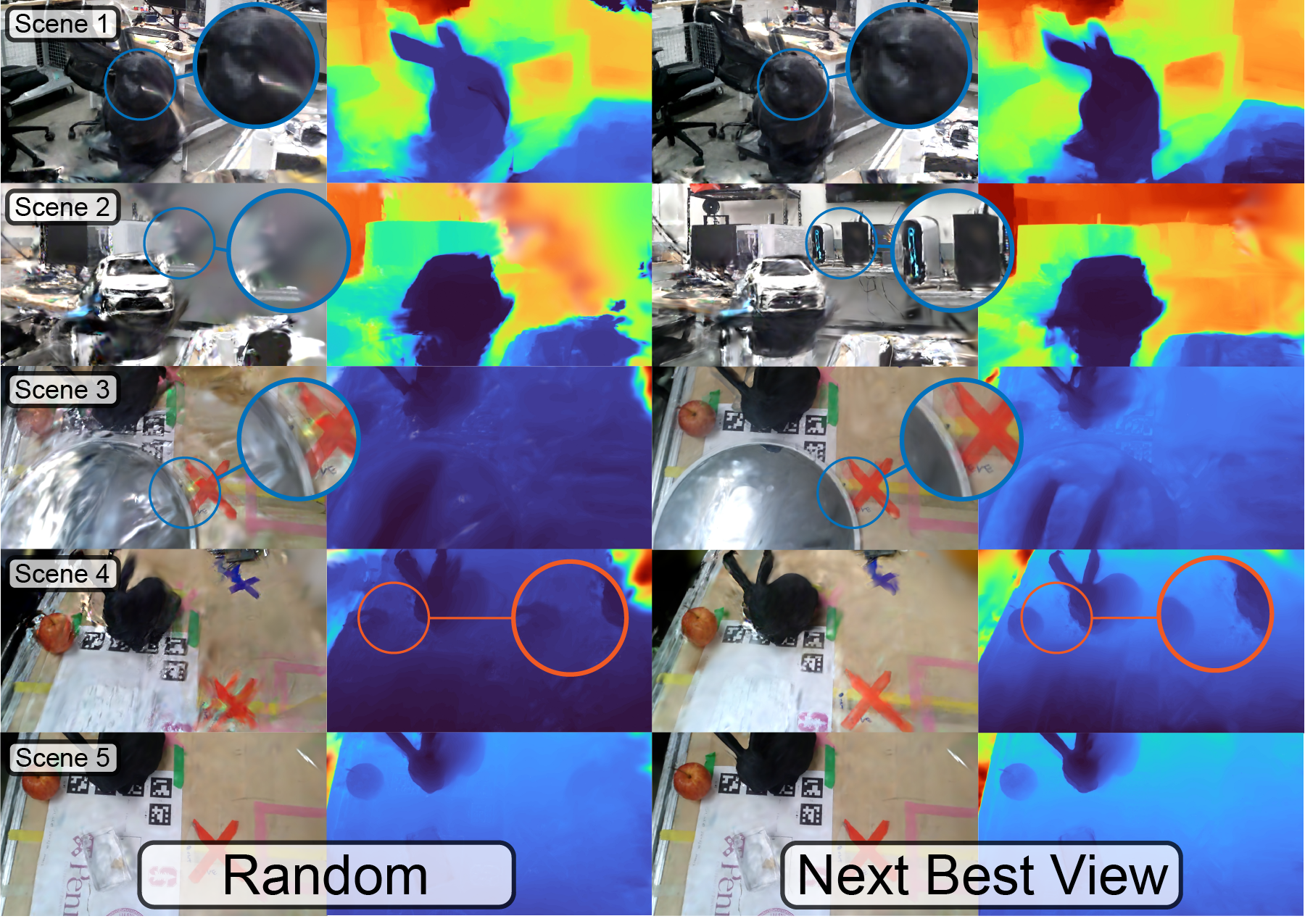}
    \caption{\textbf{Qualitative Results of Next Best View}}
    \label{fig:results}
    
\end{figure*}

\subsection{Few Shot Gaussian Splatting for Robotics}

\subsubsection{Stanford Bunny Blender} We demonstrate our improvements to Gaussian Splatting on the Touch-GS \cite{swann2024touch} dataset to highlight the improvements to vision for robotic tasks.  To start, we perform ablations starting from vanilla \gls*{3DGS}. We use the camera poses from Blender and use the perturbed ground truth depths with noise that quadratically increases with distance to simulate real depth. We use Nerfstudio \cite{nerfstudio} and implement depth supervision, normal supervision, and initialization of Gaussians from depth maps. We train on only $6$ views for 15000 steps as convergence is reached early. We test PSNR, SSIM, and LPIPs for 34 unseen novel views on 3 trials and report the mean, and additionally the mean absolute \textit{error for ground truth depth}.  

As seen in Table \ref{tab:ablation-results}, depth supervision quantitatively improves the geometric and visual quality of \gls*{3DGS}. We observed that lifting the aligned Gaussians improved the density of the scene, but was not as geometrically accurate as using SAM2 aligned depths for initialization. SAM2 aligned depths improve the visual quality across test scenes besides LPIPS (shown as the green value). Further, the mean absolute error in depth in the SAM2 aligned depths from Table \ref{tab:depth-results} is the best, highlighting its usefulness as a depth prior.

\subsubsection{Stanford Bunny Real and Challenge Objects} With a 3D printed model of the Stanford bunny, we highlight our improvements in a real scene with low resolution visual and depth data on 8 views. In Table \ref{tab:bunnyrealresults}, metric depth loss on aligned depths becomes infeasible, and our method outperforms current baselines, even Touch-GS \cite{swann2024touch}, which uses hundreds of extra touches on the same number of views. Unlike the Blender example, we notice a significant gain in using Pearson Depth Loss as compared to simple MSE loss for depth, which we attribute to monocular depth being globally aligned with  actual, noisy real world depth instead of the perturbed Blender depth. 

We further include camera optimization, which improves the quality of the scene as the camera pose of the robot is not optimal.
Finally, we test our method on a difficult mirror and prism on 8 views, with normal supervision and camera optimization turned on, as shown in Table \ref{tab:realworldchallengeresults}. We still note minor gains even though these objects present a challenge for vision alone.

\subsection{Offline FisherRF}
We highlight the improvement of \textbf{FisherRF for depth} to our two bunny scenes, reporting the mean PSNR, SSIM, and LPIPS from five trials in Table \ref{tab:realworldresultsoffline}. We train GS for 15000 steps and select a new view every 2000 steps according to FisherRF, turning camera optimization off to see the direct effect of an added view and its pose. We verify on random, FisherRF depth, RGB (baseline), and a combination of the two ($\alpha = 0.1$, $\beta = 1$) with our formulation in Eq. \ref{eq:fisher-combo}. Selecting random views performs the worst, and while the baseline FisherRF on color is a significant improvement over random views, it does not select views that improve the geometric reconstruction, as seen in Fig \ref{fig:fisher-bunny}, where the geometries to the right of the back chair are incorrect. In fact, the head of the bunny is the sharpest in FisherRF depth. We posit that FisherRF on depth additionally receives the benefits of FisherRF color -- we notice that very different colored Gaussians, which are higher areas of color uncertainty, often are associated with ambiguous depths. However, the converse is not necessarily true, which provides additional benefit to the scene. The difference is smaller between the two in the real world; however, we choose to use FisherRF on depth for the following experiments.

\subsection{Robotic FisherRF}

We apply our method to a real-world robotics scenario on challenging objects and scenes -- the Stanford bunny in two scenes, a toy Toyota Corolla, mirror, prism and 3D-printed ridged object. We use the Kinova Gen3 robot, and create an in-house pipeline in Docker using ROS, which is interfaced directly with Nerfstudio, allowing us to train a Splat and add views in real-time.

We start with only 8 random views. For each object, we keep the starting views the same for each method, and the list of candidate views every 2000 steps is kept the same (10 kinematically feasible poses). We train to 21000 steps. We set $\alpha = 0.1$, $\beta = 1$ for FisherRF combined. All experiments are run in real-time. We test each Splat on a held-out set per object of 25-40 views. The result is summarized in Table \ref{tab:realworldresults}, with qualitative results in Fig \ref{fig:results}. For all the objects, FisherRF performs better than the random baseline. We note that FisherRF on depth performs the best for the real world bunny -- better than our baseline of FisherRF on color -- which comes from FisherRF depth encouraging new views to a robot to explore the \textit{background} more, which originally is highly discontinuous. Additionally, FisherRF combined proves to be unusable compared to using pure depth. This suggests that color may sometimes hinder view selection, requesting to view areas that are simply colorful. Having found that FisherRF depth is the most reliable ablation, we compare it as our method to random view selection.  On the car, mirror, prism, and ridged objects, FisherRF performs a depth smoothing effect by suggesting new views that learn the geometry of a scene more precisely, and visually generalizes to new views. We find that real world scenes are color rich but objects are geometrically smooth.

\subsection{FisherRF and Touch Data Supervision}
\begin{figure}
    \centering
    \includegraphics[width=0.85\linewidth]{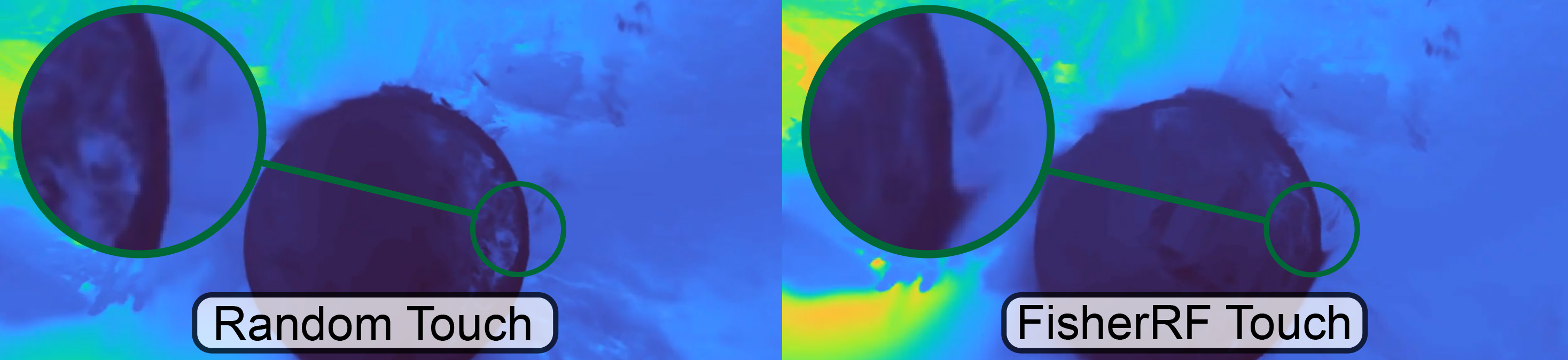}
    \caption{\textbf{FisherRF Guided Touch Selection}}
    \label{fig:touch-fisherrf}
    \vspace{-6mm}
\end{figure}

As a final experiment, we verify the usefulness of FisherRF for single \textbf{touch supervision} on a mirror example. We leverage a FisherRF-depth guided Splat from the Touch-GS mirror and perform FisherRF depth on a list of candidate touches every 100 steps until we have added 10 touches. The result is shown in Fig \ref{fig:touch-fisherrf}, where more touches are selected in the hole-like areas in the mirrors, which is due to the the holes having a higher Hessian with respect to depth.


\section{CONCLUSION}
\label{sec:conclusions}
In this work, we demonstrated the usefulness of few-shot active view and touch selection for robot manipulators using \gls*{3DGS}. We utilize SAM2 and monocular depth models to generate high quality depth maps  for Gaussian Splatting training, using FisherRF to actively select both camera and touch poses to  reconstruct an object and scene surface.

Future work entails integrating our pipeline with both vision and touch \textbf{online}. We  plan to integrate both semantic and normal uncertainty into our method, which we suggest aligns with human visual uncertainty as poorly constructed Splats look semantically confusing to the human eye. We believe that a \textbf{true} Next Best Sense will unify vision and touch beyond few-view reconstruction, as the real power in touch lies in both the reconstruction of challenging surfaces and object property identification, such as stiffness and texture.




\bibliographystyle{IEEEtran} 
\bibliography{ms} 


\end{document}

%% file: glossary.tex
\usepackage[acronym,nomain,nonumberlist]{glossaries}
\makeglossaries
\newacronym{NeRF}{NeRF}{Neural Radiance Fields}

\newacronym{3DGS}{3DGS}{3D Gaussian Splatting}
\newacronym{GS}{GS}{Gaussian Splatting}
\newacronym{SDF}{SDF}{Signed Distance Field}

\newacronym{SfM}{SfM}{Structure from Motion}

\newacronym{GPIS}{GPIS}{Gaussian Process Implicit Surface}
\newacronym{GP}{GP}{Gaussian Process}


%% file: monroe_intro.tex
As robots become more capable of performing manipulation in daily tasks, it is important for them to efficiently model the scene and relevant objects \cite{lu2024manigaussian,shorinwa2024splat}. One prominent explicit 3D representation is \gls*{3DGS} \cite{3dgs}, which reconstructs a 3D scene from RGB images and camera poses. Traditionally, the creation of \gls{3DGS} scenes is done with a). many views and b). complete human supervision. \gls{3DGS} is well-known for requiring tens of views and near-perfect camera poses; as it is prone to overfitting on sparse input views.  This approach, however, is inconsistent with many of the potential robotic applications of \gls{3DGS}. In a robotic setting, it is inefficient to collect hundreds of views, often by a human teleoperator, required for training a \gls{3DGS} model in robotic environments. For future applications of \gls{3DGS} on robotics systems, robust few-view training methods are needed. Even the assistance of depth priors in prior work on few-shot Gaussian Splatting \cite{chung2024depthregularized,paliwal2024coherentgssparsenovelview,dust3r2023, zhu2023fsgs} often relies on high quality visual data for constructing a precise 3D scene, assumes that structure from motion points are already available, and require strong visual priors. Robots with noisy depth and low resolution RGB cameras cannot benefit from this.

\begin{figure}[t]
    \centering
    \includegraphics[width=.45\textwidth]{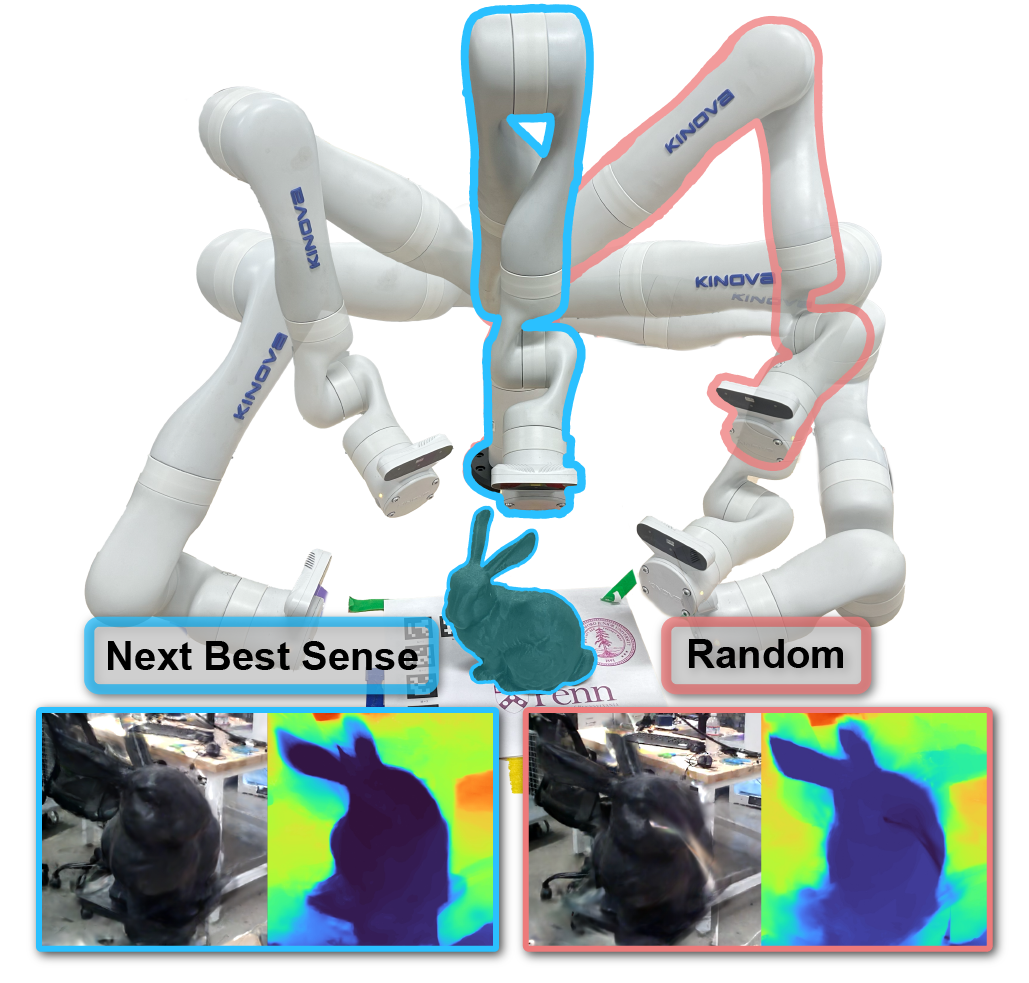}
    \caption{Our method outperforms random view selection for few view \gls{3DGS} scenes. We show a series of robot views, with the next view proposed by our method compared to random..}
    \label{fig:showcase}
    \vspace{-6mm}
\end{figure}

To address this, we propose the use of \textit{semantic depth alignment} to enhance the impact of each additional view. We draw from the power of state-of-the-art monocular depth estimation models, and with an off-the-shelf depth camera and color image, we perform a semantic depth alignment by aligning objects and the background of an image with Segment Anything Model 2 (SAM2) \cite{ravi2024sam2segmentimages}. This is used as a form of initialization in our few-shot scene. During training, depth supervision is enforced with a relaxed relative loss, which mitigates the scale ambiguities between the true depth and estimated depth from the depth estimation models. Gentle normal supervision and camera optimization guide the Gaussian Splat to reconstruct a visually accurate scene.

In addition to maximizing the impact of each of the limited training views, we propose a method to optimally select the \emph{next best view} based on depth uncertainty and optionally color uncertainty. When there are a limited number of views available, uncertainty emerges from depth discontinuities, occluded regions, and sources of aleatoric error. It is critical to understand where \textit{next} best views and touches should be made in a scene to reduce this uncertainty and construct precise representations for effective robot operation. 

In order to determine the optimal next view in a radiance field, we develop a uncertainty metric and maximize uncertainty gain. Past work in uncertainty quantification in radiance fields includes per-pixel uncertainty, ray entropy, ensemble methods, and interpreting the radiance field as a probability distribution  \cite{lin2022active, pan2022activenerf, lee2022uncertainty, shen2024estimating}. These works require significant modifications to the original radiance field architecture, become computationally infeasible in the case of ensemble-based uncertainty, and make assumptions about ray density distributions that are not generalizable. FisherRF~\cite{jiang2023fisherrf}, a recent state-of-the-art work in uncertainty quantification, uses the \textbf{Fisher Information} to quantify observed information without any ground truth data. Only requiring one backward pass, FisherRF is broadly generalizable to explicit radiance fields like Gaussian Splatting, making it a suitable method for robotics applications.


FisherRF computes the expected information gain on candidate views of \gls{3DGS} to then select the next best view. Unlike FisherRF, recent research on NBV in offline and online applications \cite{lee2022uncertainty, shen2024estimating, pan2022activenerf} leverages prior methods for uncertainty quantification and is only applied to implicit radiance fields methods like Neural Radiance Fields (NeRFs).

We extend FisherRF to select views \textit{and} touches based on \textbf{depth uncertainty}. We posit that the geometry and visual quality of a scene can be guided through depth, as we observe that regions of erroneous color in Gaussian Splats are often associated with ambiguous depth. This framework can be easily extended to touch, which has been shown to improve \gls{3DGS} quality \cite{swann2024touch}.

The integration of touch into \gls{3DGS} is still limited, however -- only  Touch-GS \cite{swann2024touch} and Snap-It, Tap-It, Splat-It \cite{comi2024snap} address touch-informed \gls{3DGS}, as compared to several works fusing vision and touch for manipulation \cite{smith2021active, suresh2023neural,yang2024binding, fu2024touch,watkins2019multi, smith20203d}. While Touch-GS demonstrates the improvements on \gls{3DGS} scenes with touch, it suffers from two shortcomings: 1. it requires a human to move the robot to desired vision and touch poses and 2. requires the order of 100-500 touches to construct a visually and geometrically sufficient scene. Snap-It, Tap-It, Splat-It similarly requires manual views. Our method automonously selects views and touches guided by uncertainty, in a two stage process, where vision coarsely reconstructs the object, and then touch refines it.

Altogether, we construct a framework that enables for the autonomous sense selection with a robotic manipulator, which we call \textit{Next Best Sense}.

\noindent \textbf{Key Contributions.} We propose \textit{Next Best Sense}, the first approach that enables uncertainty-guided view \textit{and} touch selection for training a \gls*{3DGS} for robotic manipulators. Our main contributions in Next Best Sense are as follows:
\textbf{1)} We propose an improved method in few-shot \gls{3DGS} scenes with a novel depth alignment method using Segment Anything Model 2.
\textbf{2) }We present a novel extension of FisherRF by leveraging depth uncertainty to assist in view and touch selection.
\textbf{3)} We present a closed-loop framework coupling both perception and action for training few-view scenes for robotic manipulators with FisherRF. To the best of our knowledge, this is the first work that enables next best view planning for Gaussian Splatting for robot manipulators in real-time. 
\textbf{4)} We extend our method to uncertainty-guided touch, demonstrating qualitative improvements with only 10 touches.